\documentclass{article}
\usepackage{harvard}
\usepackage{latexsym}

\newcommand{\eclipse}{ECL$^i$PS$^e$}

\newcounter{rulecnt}
\newcommand{\ES}{\mbox{$\emptyset$}}

\newcommand{\ra}{\mbox{$\:\rightarrow\:$}}

\newcommand{\La}{\mbox{$\:\Leftarrow\:$}}
\newcommand{\Ra}{\mbox{$\:\Rightarrow\:$}}

\newcommand{\A}{\mbox{$\ \wedge\ $}}

\newcommand{\Or}{\mbox{$\ \vee\ $}}

\newcommand{\sse}{\mbox{$\:\subseteq\:$}}

\newcommand{\te}{\mbox{$\exists$}}

\newcommand{\LL}{\mbox{$\ldots$}}

\newcommand{\C}[1]{\mbox{$\{{#1}\}$}}           

\newcommand{\NI}{\noindent}
\newcommand{\HB}{\hfill{$\Box$}}
\newcommand{\VV}{\vspace{5 mm}}
\newcommand{\III}{\vspace{3 mm}}
\newcommand{\II}{\vspace{2 mm}}

\newcommand{\szkew}[1]{\relax \setbox0=\hbox{\kern -24pt $\displaystyle#1$\kern 0pt }%
\box0}
{\catcode`\@=11 \global\let\ifjusthvtest@=\iffalse}

\newcounter{oldmycaption}

\newcommand{\p}[2]{\langle #1 \ ; \ #2 \rangle}

\newcommand{\Proof}{\NI
                    {\bf Proof.}\ }
\newtheorem{theorem}{Theorem}[section]
\newtheorem{defined}[theorem]{Definition}
\newenvironment{definition}{\begin{defined} \rm}{\end{defined}}
\newtheorem{exa}[theorem]{Example}

\newtheorem{exe}{Exercise}

\newtheorem{pro}{Problem}

\newcounter{symbol}
\newcommand{\indexsyma}[1]%
{\stepcounter{symbol}\index{zzz1 \thesymbol @\protect#1}}
\newcommand{\indexsymb}[1]%
{\stepcounter{symbol}\index{zzz2 \thesymbol @\protect#1}}
\newcommand{\indexsymc}[1]%
{\stepcounter{symbol}\index{zzz3 \thesymbol @\protect#1}}
\newcommand{\indexsymd}[1]%
{\stepcounter{symbol}\index{zzz4 \thesymbol @\protect#1}}
\newcommand{\indexsyme}[1]%
{\stepcounter{symbol}\index{zzz5 \thesymbol @\protect#1}}

\normalbaselines

\begin{document}

\pagestyle{plain}

\author{Krzysztof R. Apt \\
        {\em CWI } \\
        {\em P.O. Box 94079, 1090 GB Amsterdam, The Netherlands} \\
        and \\
        {\em University of Amsterdam, The Netherlands }
}
\title{Some Remarks on Boolean Constraint Propagation}
\date{}
\maketitle

\begin{abstract}
  We study here the well-known propagation rules for Boolean
  constraints.  First we propose a simple notion of completeness for
  sets of such rules and establish a completeness result. Then we show
  an equivalence in an appropriate sense between Boolean constraint
  propagation and unit propagation, a form of resolution for
  propositional logic.

  Subsequently we characterize one set of such rules by means of the
  notion of hyper-arc consistency introduced in \citeasnoun{MM88}.
  Also, we clarify the status of a similar, though different, set of
  rules introduced in \citeasnoun{simonis89a} and more fully in
  \citeasnoun{CD96}.

\end{abstract}

\section{Introduction}

\subsection{Motivation}
\label{subsec:motivation}

Boolean constraints form a special case of constraint satisfaction
problems in which the constraints are defined by means of Boolean
formulas. The most common representation uses basic constraints that
represent the typical connectives, such as {\em and, not\/} etc.

To reason about Boolean constraints one often uses rules such as:

\begin{equation}
\mbox{``for $x \A y = z$, if $z$ holds, then both $x$ and $y$ hold''}
\label{equ:rule1}
\end{equation}
\noindent
or

\begin{equation}
\mbox{``for $x \Or y = z$, if $x$ does not hold, then $y = z$ holds.''}
\label{equ:rule2}
\end{equation}

These rules allow us to propagate the information that some values in
a Boolean constraint are known.  This type of inferences have been
used for a long time. In \citeasnoun{McA80} they are
explained informally; in \citeasnoun{McA90} they are called {\em
  Boolean constraint propagation}.  In \citeasnoun{simonis89a} such
rules are formulated explicitly and used to propagate known values
through the circuit when generating tests for combinatorial circuits.
More recently, these rules were used in \citeasnoun{CD96} as a basis
for an efficient implementation of a Boolean constraint solver.

In this paper we put together various simple observations concerning
Boolean constraint propagation.  The main difficulty lies in a proper
setting up of the framework. Once this is done the results easily
follow.

To start with, in Section \ref{sec:completeness}, we answer the
question in what sense a set of such rules can be complete. To this
end we introduce a notion of completeness based on the notions of
minimal rules and valid rules and show completeness for one set
of such rules.  In Section \ref{sec:unit} we relate Boolean constraint
propagation to unit propagation, a form of resolution for
propositional logic, by explaining in what sense each method 
can be simulated by the other.

Next, in Section \ref{sec:proof-theoretic} we introduce proof rules
that act on CSP's. This allows us to provide in Section
\ref{sec:hyper-arc} an alternative characterization for one set of
rules by means of the notion of hyper-arc consistency of
\citeasnoun{MM88} (we use here the terminology of \citeasnoun{MS98b}).
In Section \ref{sec:cd} we clarify the status of another, more
commonly used, set of such rules given for the {\em and\/} constraint
in \citeasnoun{simonis89a} and for other connectives in
\citeasnoun{CD96}.  In the final
section we relate Boolean constraint propagation to the {\tt CHR}
language of \citeasnoun{fruhwirth-constraint-95}.

\subsection{Preliminaries}

We review here the notions used in the sequel.

Consider a finite sequence of variables $Y := y_1, \LL, y_k$ where $k
\geq 0$, with respective domains ${\cal D} := D_1, \LL, D_k$
associated with them.  So each variable $y_i$ ranges over the domain
$D_i$.  By a {\em constraint} $C$ on $Y$ we mean a subset of $D_1
\times \LL \times D_k$.  If $C$ equals $D_1 \times \LL \times D_k$,
then we say that $C$ is {\em solved}.

Now, by a {\em constraint satisfaction problem}, CSP in short, we mean a
finite sequence of variables $X:= x_1, \LL, x_n$ with respective
domains ${\cal D} := D_1, \LL, D_n$, together with a finite set $\cal
C$ of constraints, each on a subsequence of $X$.  We write such a CSP
as $\p{{\cal C}}{{\cal DE}}$, where ${\cal DE} := x_1 \in D_1, \LL,
x_n \in D_n$ and call each construct of the form $x \in D$ a {\em
domain expression}.  To simplify the notation from now on we omit the
``\{ \}'' brackets when presenting specific sets of constraints ${\cal
C}$.

Consider now an element $d := d_1, \LL, d_n$ of $D_1 \times \LL \times
D_n$ and a subsequence $Y := x_{i_1}, \LL, x_{i_\ell}$ of $X$. Then we
denote by $d[Y]$ the sequence $d_{i_1}, \LL, d_{i_{\ell}}$.  By the
{\em domain of $Y$\/} we mean the set of all tuples from $D_{i_1}
\times \cdots \times D_{i_\ell}$.
By a {\em solution\/} to  $\p{{\cal C}}{x_1 \in D_1, \LL, x_n \in D_n}$
we mean an element $d \in D_1 \times \LL \times D_n$ such that for
each constraint $C \in {\cal C}$ on a sequence of variables $X$
we have $d[X] \in C$.

Next, we call a CSP {\em failed\/} if some of its domains is empty.
Given two CSP's $\phi$ and $\psi$, we call $\phi$ a {\em reformulation\/} of
$\psi$ if the removal of solved constraints from $\phi$ and $\psi$
yields the same CSP.  We call two CSP's with the same sequence of
variables {\em equivalent\/} if they have the same set of solutions.
Clearly, two CSP's such that one is a reformulation of another are equivalent.

Finally, given a constraint $c$ on the variables $x_1, \LL, x_n$
with respective domains $D_1, \LL, D_n$, and a sequence of 
domains $D'_1, \LL, D'_n$ such that 
for $i \in [1..n]$ we have $D'_i \sse D_i$, we say that
$c'$ equals $c$ {\em restricted to the domains $D'_1, \LL, D'_n$} if
$c' = c \cap (D'_1 \times \dots \times D'_n)$.

In this paper we focus on Boolean constraint satisfaction problems.
They deal with Boolean variables and constraints on them defined by
means of Boolean connectives and equality.  Let us introduce the
relevant definitions.

By a {\em Boolean variable\/} we mean a variable which ranges over  the
domain which consists of two values: $0$ denoting {\bf false} and
$1$ denoting {\bf true}.
By a {\em Boolean domain expression\/} we mean an expression of the form
$x \in D$ where $D \sse \C{0,1}$.
In what follows we write the Boolean domain expression $x \in \C{1}$ as
$x = 1$ and $x \in \C{0}$ as $x = 0$.

In the sequel $x,y,z$ denote different Boolean variables.
We distinguish four Boolean constraints:

\begin{itemize}

\item $x = y$; we call it the  {\em equality\/} constraint,

\item $\neg x = y$; we call it the {\em NOT\/} constraint,

\item $x \A y = z$; we call it the {\em AND\/} constraint,

\item $x \Or y = z$; we call it the {\em OR\/} constraint,

\end{itemize}
and interpret them in the expected way.

Finally, by a {\em Boolean constraint satisfaction problem\/}, in short
{\em Boolean CSP\/},  we mean a CSP with Boolean domain expressions and each
constraint of which is a  Boolean constraint restricted to the adopted domains.

For example, the Boolean CSP 

\[
\p{x \A y = z, \neg x = y}{x = 1, y \in \C{0,1}, z \in \C{0,1}}
\]
can be alternatively written as
\[
\p{C_1, C_2}{x = 1, y \in \C{0,1}, z \in \C{0,1}},
\]
where $C_1 = \C{(1,1,1), (1,0,0}$ is a constraint on $x,y,z$
and $C_2 = \C{(1,0)}$ is a constraint on $x,y$.

In this paper we shall relate Boolean constraints to clauses as used
in the resolution method.  The relevant notions are the following
ones.

A {\em literal\/} is a Boolean variable or its negation; a {\em
clause\/} is a (possibly empty) disjunction of different literals.  We
denote the complement of the literal $u$ by $\bar{u}$.  A clause with
a single literal is called a {\em unit clause}.  We write $u \Or Q$ to
denote a clause that contains the literal $u$ as a disjunct; $Q$ is
the disjunction of the remaining literals.

\section{The Proof System {\em BOOL} and Its Completeness}
\label{sec:completeness}
The rules such as the ones given in Section \ref{subsec:motivation}
can be naturally interpreted as implications over the constraint
formed by the truth table of the connective in question.  For
instance, rule (\ref{equ:rule1}) can be viewed as the implication
\[
z = 1 \ra x = 1, y = 1
\]
over the {\em AND\/} constraint on the variables $x,y,z$
determined by the table:
\III

\begin{center}
\begin{tabular}{|l|l|l|}
\hline
$x$ & $y$ & $z$ \\ \hline \hline
0 & 0 & 0 \\ 
0 & 1 & 0 \\ 
1 & 0 & 0 \\ 
1 & 1 & 1 \\ \hline
\end{tabular}
\end{center}

With this interpretation ``completeness'' of a set of such rules
can be naturally interpreted as the question whether the set implies
all other valid rules.
These concepts can be made precise as follows
(see essentially \citeasnoun{AM99}).

\begin{definition}
  Consider a constraint $C$ on a sequence of variables {\it VAR\/},
  two disjoint non-empty subsequences $X$ and $Y$ of {\it VAR\/}, a
  tuple $s$ of elements from the domain of $X$ and a tuple $t$ of
  elements from the domain of $Y$.  We call $X = s \ra Y = t$ a {\em
    rule\/} ({\em for\/} $C$).

\begin{itemize}
\item 
We say that 
$X = s \ra Y = t$ is {\em valid\/} ({\em for\/} $C$)
if for every tuple $d \in C$ the equality $d[X] = s$ implies 
the equality $d[Y] = t$.

\item We say that 
$X = s \ra Y = t$ is {\em feasible\/} ({\em for\/} $C$)
if for some tuple $d \in C$ the equality $d[X] = s$ holds.
\end{itemize}

Suppose that a sequence of variables $Z$ extends $X$ and a tuple of
elements $u$ from the domain of $Z$ extends $s$.  We say then that $Z
= u$ {\em extends\/} $X = s$.  We now say that the rule $Z = u \ra U = v$
{\em is implied\/} by the rule $X = s \ra Y = t$ if $Z = u$ extends $X
= s$ and $Y = t$ extends $U = v$.

We call a rule {\em minimal\/} if it is feasible
and is not properly implied by a valid rule.
Finally, we call a set of rules ${\cal R}$ for a constraint $C$
{\em complete\/} if it consists of all minimal valid rules for $C$.
\HB 
\end{definition}

Take for example the {\em AND\/} constraint. The 
rule $z = 1 \ra y = 1$ is implied by the rule $z = 1 \ra x = 1, y = 1$.
Since both of them are valid, the former rule is not minimal.
Both rules are feasible, whereas the rule $z = 0, x = 1 \ra y = 0$ is not.
One can check that the rule $z = 1 \ra x = 1, y = 1$ is minimal.

Consider now the set of rules presented in Table \ref{tab:bool}, where
for the sake of clarity we attached to each implication the Boolean
constraint in question.
Call the resulting proof system {\em BOOL}.

\begin{table}[htbp]
    \leavevmode
\[
\begin{array}{|ll|}  \hline
{\it EQU \ 1} & x = y, x = 1 \ra y = 1 \\
{\it EQU \ 2} & x = y, y = 1 \ra x = 1 \\
{\it EQU \ 3} & x = y, x = 0 \ra y = 0 \\
{\it EQU \ 4} & x = y, y = 0 \ra x = 0 \\[2mm]
{\it NOT \ 1} &\neg x = y, x = 1 \ra y = 0 \\
{\it NOT \ 2} &\neg x = y, x = 0 \ra y = 1 \\
{\it NOT \ 3} &\neg x = y, y = 1 \ra x = 0 \\
{\it NOT \ 4} &\neg x = y, y = 0 \ra x = 1 \\[2mm]
{\it AND \ 1} & x \A y = z, x = 1, y = 1 \ra z = 1 \\
{\it AND \ 2} & x \A y = z, x = 1, z = 0 \ra y = 0 \\
{\it AND \ 3} & x \A y = z, y = 1, z = 0 \ra x = 0 \\
{\it AND \ 4} & x \A y = z, x = 0 \ra z = 0 \\
{\it AND \ 5} & x \A y = z, y = 0 \ra z = 0 \\
{\it AND \ 6} & x \A y = z, z = 1 \ra x = 1, y = 1 \\[2mm]
{\it OR \ 1}  & x \Or y = z, x = 1 \ra z = 1 \\
{\it OR \ 2}  & x \Or y = z, x = 0, y = 0 \ra z = 0 \\
{\it OR \ 3}  & x \Or y = z, x = 0, z = 1 \ra y = 1 \\
{\it OR \ 4}  & x \Or y = z, y = 0, z = 1 \ra x = 1 \\
{\it OR \ 5}  & x \Or y = z, y = 1 \ra z = 1 \\
{\it OR \ 6}  & x \Or y = z, z = 0 \ra x = 0, y = 0 \\ \hline
\end{array}
\]

\caption{Proof system {\em BOOL}}
\label{tab:bool}
\end{table}

A natural question arises whether some rules have been omitted in the
proof system {\it BOOL\/}.  Observe for example that no rule is
introduced for $x \A y = z$ when $z = 0$.  In this case either $x = 0$
or $y = 0$ holds, but $x = 0 \Or y = 0$ is not a legal conclusion of a
rule.  Alternatively, either $x = z$ or $y = z$ holds, but $x = z \Or
y = z$ is not a legal conclusion of a rule either.  The same
considerations apply to $x \Or y = z$ when $z = 1$.

Also, we noted already that rule {\em AND \ 6} corresponds to rule
(\ref{equ:rule1}). In contrast, no rule corresponds to rule
(\ref{equ:rule2}). The following simple result clarifies the situation.

\begin{theorem}[Completeness] \label{thm:Bcharacterization}
For each Boolean constraint the corresponding set of rules
given in the proof system {\it BOOL\/} is complete.
\end{theorem}
\Proof 
The claim follows by a straightforward exhaustive analysis of the
valid rules for each considered Boolean constraint. Clearly, such an
argument can be mechanized by generating all minimal rules for each
Boolean constraint. This was done in \citeasnoun{AM99} for the case of
arbitrary finite constraints and rules of the form $X = s \ra Y \neq
t$ that have an obvious interpretation.  Now, for the case of Boolean
constraints each domain has two elements, so each rule of the form $X
= s \ra Y \neq t$ has a ``dual'' of the form $X = s \ra Y
= t'$, where $t'$ is obtained from $t$ by a bitwise complement.
\HB 
\VV

It is useful to mention that the abovementioned program, implemented
in \eclipse{}, generated the appropriate rules for the {\em AND\/}
constraint in 0.02 seconds and similarly for the other three Boolean
constraints.

\section{Relation to Unit Propagation}
\label{sec:unit}

The considerations of the previous section clarify the matter of
completeness.  We still should explain how the rules of the proof
system {\em BOOL\/} are supposed to be applied.  To this end we
consider finite sets of Boolean constraints and literals and interpret
the rules as proof rules applied to
such sets.  We illustrate it by means of an example.

Consider {\em OR 3} rule.
We interpret it as the following proof rule:
\[
\frac
{x \Or y = z, \neg x , z}
{\neg x , y, z}
\]

We define now the result of applying a rule of {\em BOOL\/} to a
finite set of Boolean constraints and literals as expected: an
application of the rule results in the replacement of (the subset
corresponding to) the premise by (the subset corresponding to) the
conclusion.  This interpretation of the rules of {\em BOOL\/} allows
us to derive conclusions that coincide with the informal use of such
rules. In the case of {\em OR 3} rule the constraint $x \Or y = z$ is
dropped as no other inference using it can be made, while the literal
$z$ is retained as other inferences using it are still possible.

In this section we relate so interpreted proof system {\em BOOL\/} to
unit propagation, a form of propositional resolution (see,
e.g. \citeasnoun{ZS96}) that is a component of the Davis-Putnam
algorithm for the satisfiability problem (see \citeasnoun{DP60}).

We consider two types of operations on a set of clauses:

\begin{itemize}

\item {\em unit resolution\/} (w.r.t. the literal $u$): given a unit
  clause $u$ and a clause $\bar{u} \Or Q$ replace $\bar{u} \Or Q$
  by $Q$,

\item {\em unit subsumption\/} (w.r.t. the literal $u$): given a unit
  clause $u$ and a clause $u \Or Q$ delete $u \Or Q$.

\end{itemize}
By {\em unit propagation\/} we mean one of the above two operations.

We now translate each Boolean constraint to  a set of clauses as follows.
We replace
\begin{itemize}

\item each equality constraint $x = y$
by the clauses
$x \Or \neg y, \neg x \Or y$,

\item each {\em NOT\/} constraint $\neg x = y$
by the clauses 
$x \Or y, \neg x \Or \neg y$,

\item each {\em AND\/} constraint $x \A y = z$
by the clauses 
$\neg x \Or \neg y \Or z, x \Or \neg z, y \Or \neg z$,

\item each {\em OR\/} constraint $x \Or y = z$
by the clauses 
$\neg x \Or z, \neg y \Or z, x \Or y \Or \neg z$.

\end{itemize}

Given a finite set of Boolean constraints and literals ${\cal S}$
we denote by $\phi_{{\cal S}}$ the resulting translation of this set
into a set of clauses.
It is straightforward to see that this translation maintains
equivalence.  

In what follows, given two sets of Boolean constraints and literals ${\cal S}_1$
and ${\cal S}_2$, we write ${\cal S}_1 \vdash_{BOOL} {\cal S}_2$ to denote the
fact that ${\cal S}_2$ is obtained by a single application of
a rule of the {\it BOOL\/} system to ${\cal S}_1$,
and  ${\cal S}_1 \vdash^{\leq i}_{BOOL} {\cal S}_2$ to denote the
fact that ${\cal S}_2$ is obtained by up to $i$ applications of the
rules of the {\it BOOL\/} system to ${\cal S}_1$.

Analogously, given two sets of clauses $\phi_1$ and $\phi_2$, we write
${\phi}_1 \vdash_{UNIT} {\phi}_2$ to denote the
fact that ${\phi}_2$ is obtained by a single application of the
unit propagation to  ${\phi}_1$, and ${\phi}_1 \vdash^{\leq i}_{UNIT} {\phi}_2$ to denote the
fact that ${\phi}_2$ is obtained by up to $i$ applications of the
unit propagation to  ${\phi}_1$.

The following result relates the proof system {\em
  BOOL\/} to unit propagation.

\begin{theorem}[Reduction 1] \label{thm:reduction1}
  Consider two finite sets of Boolean constraints and literals ${\cal S}_1$
  and ${\cal S}_2$.  Suppose that ${\cal S}_1 \vdash_{BOOL} {\cal S}_2$.
Then $\phi_{{\cal S}_1} \vdash^{\leq 4}_{UNIT} \phi_{{\cal S}_2}$.
\end{theorem}
\Proof We need to analyze each of the 20 rules of {\it BOOL\/}. We
illustrate the argument on one, arbitrary selected rule, {\em OR 3}.

Suppose that ${\cal S}_2$ is the result of applying rule {\em OR 3\/}
to ${\cal S}_1$. Recall that this rule is interpreted as
\[
\frac
{x \Or y = z, \neg x , z}
{\neg x , y, z}
\]
The assumption of this rule
translates to the following set of clauses:
\[
\C{\neg x \Or z, \neg y \Or z, x \Or y \Or \neg z, \neg x , z}.
\]
By the application of the unit resolution w.r.t. $z$ we obtain the set 
\[
\C{\neg x \Or z, \neg y \Or z, x \Or y, \neg x , z},
\]
from which by two applications of the unit subsumption w.r.t. $z$ we
obtain the set
\[
\C{x \Or y, \neg x , z}.
\]
By the application of the unit resolution w.r.t. $\neg x$ we now obtain
the set 
\[
\C{\neg x, y, z}
\] 
which corresponds to the conclusion of rule {\em OR 3}.

For other rules the argument is equally straightforward.
\HB
\VV

The converse relation is a bit more complicated since to
translate clauses to Boolean constraints we need to use
auxiliary variables. 
First, we translate each expression of
the form $Q = z$, where $Q$ is a clause and $z$ a variable, to
a finite set of Boolean constraints and literals.
We proceed by induction on the number of literals in $Q$.

If $Q$ is a unit clause, then $Q = z$ is either an equality constraint
or a {\em NOT} constraint and we put $trans(Q = z) := \C{Q = z}$.
Otherwise $Q$ is of the form $u \Or Q_1$
and we define
\[
trans(x \Or Q_1 = z) := \C{x \Or y = z} \cup trans(Q_1 = y),
\]
where $y$ is a fresh variable,
\[
trans(\neg x \Or Q_1 = z) := \C{\neg x = v, v \Or y = z} \cup trans(Q_1 = y),
\]
where $v$ and $y$ are fresh variables.

Finally, we put for a unit clause $u$
\[
trans(u) := \C{u},
\]
and for a non-unit clause $Q$
\[
trans(Q) := \C{z} \cup trans(Q = z),
\]
where $z$ is a fresh variable.

Note that for a non-unit clause $Q$ the resulting finite set of Boolean
constraints and literals $trans(Q)$ depends on the order in which the
literals of $Q$ are selected and on the specific choice of the fresh
variables, so it is not uniquely determined.
However, it is clear that for each such translation
$trans(Q)$, the clause $Q$ is equivalent to $\te \bar{z} \: trans(Q)$,
where $\bar{z}$ is the sequence of the introduced fresh variables.

Given now a finite set of clauses ${\phi}$ we translate each of its
clauses separately and call thus obtained finite set of Boolean
constraints and literals a {\em translation of ${\phi}$ to a 
finite set of Boolean constraints and literals}.

Below, given two sets of Boolean constraints and literals ${\cal C}$
and ${\cal S}$ we say that ${\cal C}$ {\em semantically follows from a
set ${\cal S}$\/} if every valuation that satisfies ${\cal S}$ can be
extended to a valuation that satisfies ${\cal C}$.

We then have the following result.

\begin{theorem}[Reduction 2] \label{thm:reduction2}
  Consider two finite sets of clauses ${\phi}_1$ and ${\phi}_2$.
  Suppose that ${\phi}_1 \vdash_{UNIT} {\phi}_2$. Then for some
  translations ${\cal S}_1$ and ${\cal S}_2$ of ${\phi}_1$ and
  ${\phi}_2$ to finite sets of Boolean constraints and literals and
  some set of Boolean constraints and literals ${\cal C}$ we have
  ${\cal S}_1 \vdash^{\leq 3}_{BOOL} {\cal S}_2 \cup {\cal C}$, where
  ${\cal C}$ semantically follows from ${\cal S}_2$.

\end{theorem}
Informally, the reduction from ${\cal S}_1$ to ${\cal S}_2$ yields
additionally some redundant set of Boolean constraints and literals
${\cal C}$.
\NI

\Proof 
Consider first the unit resolution. It leads to a replacement of 
$\bar{u} \Or Q$ by $Q$ in presence of the unit clause $u$.
Suppose that $u$ is a Boolean variable $x$. Then 
$\bar{u}$ is $\neg x$. 

We now have for some fresh variables $v,y$ and $z$
\[
trans(\neg x \Or Q) = \C{z} \cup \C{\neg x = v, v \Or y = z} \cup trans(Q = y).
\]
So the clauses $x$ and $\neg x \Or Q$ translate to the set of
Boolean constraints and literals
\[
\C{x, z, \neg x = v, v \Or y = z} \cup trans(Q = y).
\]
By the application of the {\em NOT 1\/} rule we now obtain the set
\[
\C{x, z, \neg v, v \Or y = z} \cup trans(Q = y),
\]
from which by the {\em OR 3\/} rule we obtain 
\begin{equation}
\C{x, z, \neg v, y} \cup trans(Q = y).
\label{equ:trans}
\end{equation}

Now, if $Q$ is a unit clause, then the set (\ref{equ:trans})
equals
\[
\C{x, z, \neg v, y, Q = y}
\]
from which we get by the {\em EQU 2\/} rule
\[
\C{x, z, \neg v, y, Q},
\]
i.e., the set $trans(x) \cup trans(Q) \cup \C{z, \neg v, y}$.  Since
$v,y$ and $z$ are fresh, $\C{z, \neg v, y}$ semantically follows from
$trans(x) \cup trans(Q)$.

If $Q$ is a non-unit clause we can assume that
\[
trans(Q) = \C{y} \cup trans(Q = y),
\]
so the set (\ref{equ:trans}) equals $trans(x) \cup trans(Q) \cup \C{z,\neg v}$.  
Since $v$ and $z$ are fresh, $\C{z, \neg v}$ semantically
follows from $trans(x) \cup trans(Q)$.

The argument in case $u$ is negation of a Boolean variable is
even more straightforward.
\II

Consider now the unit subsumption.  It leads to a deletion of the clause
$u \Or Q$ in presence of the unit clause $u$.  Suppose that $u$ is
$\neg x$ for some Boolean variable $x$. We have
for some fresh variables $v,y$ and $z$
\[
trans(\neg x \Or Q) = \C{z, \neg x = v, v \Or y = z} \cup trans(Q = y).
\]
So the clauses $\neg x$ and $\neg x \Or Q$ translate to the set of
Boolean constraints and literals
\[
\C{\neg x, z, \neg x = v, v \Or y = z} \cup trans(Q = y).
\]
By the application of the {\em NOT 2\/} rule we now obtain the set
\[
\C{\neg x, z, v, v \Or y = z} \cup trans(Q = y),
\]
from which by the {\em OR 1\/} rule we obtain
\[
\C{\neg x, z, v} \cup trans(Q = y).
\]

It is now easy to see that the set $\C{z, v} \cup trans(Q = y)$
semantically follows from $\C{\neg x}$.
Indeed, a straightforward proof by induction shows that for any clause $Q$ the set 
$trans(Q = y)$ is satisfiable.  
\HB
\VV

The above two results clarify the relationship between Boolean
constraint propagation and unit propagation. They show that each
method can be simulated by another in constant time, albeit the
simulation of the unit propagation by means of the Boolean constraint
propagation leads to a generation of redundant constraints.

A relation  between Boolean constraint propagation and the
Davis-Putnam algorithm was already mentioned in \citeasnoun[page
1]{McA80}, where it is stated without any further explanation that 
``propositional [i.e., Boolean] constraint propagation [...]  was
originally described, in essence, by \citeasnoun{DP60} ''. But to our
knowledge this connection was not made precise.  

\section{A Proof Theoretic Framework}
\label{sec:proof-theoretic}

We now proceed towards another characterization of 
the proof system {\em BOOL\/} in constraint processing terms.
In the previous section we considered finite sets of Boolean
constraints and literals. We now need to translate them
into Boolean CSP's by interpreting in an appropriate
way the literals belonging to such a set.

Given a Boolean variable $x$ there are four sets of literals
concerning $x$. We interpret each of them as a Boolean domain
expression, as follows:

\begin{itemize}

\item $\ES$ by $x \in \C{0,1}$,

\item $\C{x}$ by $x \in \C{1}$,

\item $\C{\neg x}$ by $x \in \C{0}$,

\item $\C{x, \neg x}$ by $x \in \ES$.

\end{itemize}

This interpretation entails a translation 
of finite sets of Boolean constraints and literals
to Boolean CSP's.
For example, the set $\C{x \Or y = z, \neg x , z}$ (that corresponds
to the premise of {\em OR 3} rule) translates to the Boolean CSP
\[
\p{x \Or y = z}{x \in \C{0}, y \in \C{0,1}, z \in \C{1}}.
\]

It is straightforward to see that this translation preserves
equivalence in the sense that $(d_1, \LL, d_n)$ is a solution to a
Boolean CSP ${\cal P} := \p{{\cal C}}{x_1 \in D_1, \LL, x_n \in D_n}$
iff the assignment $(x_1/d_1, \LL, x_n/d_n)$ satisfies the original set of
Boolean constraints and literals.

This translation also leads to another interpretation of the rules of the proof
system {\em BOOL}.  We interpreted them as rules on the finite sets of
Boolean constraints and literals. By means of the above translation
they become rules on Boolean CSP's.

Note that for a set $L$ of literals concerning $x$ 
that translates into the Boolean domain expression $x \in D$, 
the set $L \cup \C{x}$ translates into the Boolean domain expression
$x \in D \cap \C{1}$, and similarly for the literal $\neg x$.
Consequently, the rule
\[
\neg x = y, y = 0 \ra x = 1,
\]
translates into
\[
\frac
{\p{\neg x = y}{x \in D_x, y = 0}}
{\p{}{x \in D_x \cap \C{1}, y = 0}}
\]
and the rule
\[
x \A y = z, z = 1 \ra x = 1, y = 1,
\]
translates into
\[
\frac
{\p{x \A y = z}{x \in D_x, y \in D_y, z = 1}}
{\p{}{x \in D_x \cap \C{1}, y \in D_y \cap \C{1}, z = 1}}
\]
In addition the rule
\[
x \Or y = z, x = 0 \ra y = z,
\]
that naturally corresponds to rule (\ref{equ:rule2})
of in Section \ref{subsec:motivation}, translates into
\[
\frac
{\p{x \Or y = z}{x = 0, y \in D_y, z \in D_z}}
{\p{y = z}{x = 0, y \in D_y, z \in D_z}}
\]

This brings us to the proof theoretic framework introduced in
\citeasnoun{Apt98a}.  We briefly recall the relevant definitions.  The
crucial concept that we need is that of a CSP being closed under the
applications of a proof rule.
In the above paper we introduced two types of proof rules for CSP's:
{\em deterministic\/} and {\em splitting}.  Here we only use the
deterministic ones. These rules are of the form
\[
\frac{\phi}{\psi}
\]
where $\phi$ and $\psi$ are CSP's.  

Consider now a CSP of the form $\p{{\cal C} \cup {\cal C}_1}{{\cal D}
  \cup {\cal D}_1}$ and a deterministic rule of the form
\begin{equation}
\frac
{\p{{\cal C}_1}{{\cal D}_1}}
{\p{{\cal C}_2}{{\cal D}_2}}
\label{eq:det}
\end{equation}
We then say that rule (\ref{eq:det}) {\em can be applied\/} to
$\p{{\cal C}  \cup {\cal C}_1}{{\cal D} \cup {\cal D}_1}$
and call
\[
\p{{\cal C}  \cup {\cal C}_2}{{\cal D} \cup {\cal D}_2}
\]
the {\em result of applying it to} $\p{{\cal C}
  \cup {\cal C}_1}{{\cal D} \cup {\cal D}_1}$.  If $\p{{\cal C} \cup
  {\cal C}_2}{{\cal D} \cup {\cal D}_2}$ is not a reformulation of
$\p{{\cal C} \cup {\cal C}_1}{{\cal D} \cup {\cal D}_1}$, then we say
that it is the result of a {\em relevant application of rule}
(\ref{eq:det}) {\em to} $\p{{\cal C} \cup {\cal C}_1}{{\cal D} \cup
{\cal D}_1}$.

Finally, given a CSP $\phi$ and a deterministic rule $R$, we say
that $\phi$ is {\em closed under the applications of $R$\/} if
either $R$ cannot be applied to $\phi$ or no application of it to
$\phi$ is relevant.

Take for example the Boolean CSP 
$\phi := \p{x \A y = z}{x = 1, y = 0, z = 0}$.
This CSP is closed under the applications of the rule
\[
\frac
{\p{x \A y = z}{x = 1, y \in D_y, z \in D_z}}
{\p{y = z}{x = 1, y \in D_y, z \in D_z}}
\]
Indeed, this rule can be applied to $\phi$; the outcome is 
$\psi := \p{y = z}{x = 1, y = 0, z = 0}$. After the removal 
of solved constraints from $\phi$ and $\psi$ we get in both cases the
solved CSP $\p{\ES}{x = 1, y = 0, z = 0}$. 

In contrast, the Boolean CSP $\phi := \p{x \A y = z}{x = 1, y \in
\C{0,1}, z \in \C{0,1}}$ is not closed under the applications of the
above rule because $\p{y = z}{x = 1, y \in \C{0,1}, z \in \C{0,1}}$ is
not a reformulation of $\phi$.

In what follows we identify the rules of the proof system {\em BOOL\/}
with their counterparts that act on Boolean CSP's.
At this stage we introduced two interpretations of the rules of the
proof system {\em BOOL\/}: one on the finite sets of Boolean
constraints and literals and the other on Boolean CSP's.  It is
straightforward to check that these interpretations correspond in the
following sense. Consider two finite sets of Boolean constraints and
literals ${\cal S}_1$ and ${\cal S}_2$ that translate respectively to
the Boolean CSP's ${\cal P}_1$ and ${\cal P}_2$ and a rule $r$ of {\em
  BOOL}.  Then in the first interpretation $r$ transforms ${\cal S}_1$
into ${\cal S}_2$ iff in the second interpretation it transforms
${\cal P}_1$ into ${\cal P}_2$.

It is worthwhile to note that the Characterization Theorem
\ref{thm:Barc} can be proved indirectly by using the theoretical
results established in \citeasnoun{AM99} together with the output of
the already mentioned in Section \ref{sec:completeness} program that
automatically generates proof rules from the truth tables, or more
generally, from a table representing a finite constraint.

\section{Relation to Hyper-arc Consistency}
\label{sec:hyper-arc}

We now return to CSP's.  In \citeasnoun{MM88} a
generalization of the notion of arc consistency of
\citeasnoun{mackworth-consistency} from binary constraints to
arbitrary constraints was introduced. Let us recall the definition.

\begin{definition} \mbox{} \\[-6mm]

\begin{itemize}

\item A constraint $C$ is called {\em hyper-arc consistent\/}
if for every variable of it each value in its domain participates
in a solution to $C$.

\item A CSP is called {\em hyper-arc consistent\/} if every constraint of it is.
\HB
\end{itemize}
\end{definition}

The following result characterizes the proof system {\it BOOL} in
terms of the notion of hyper-arc consistency for Boolean CSP's.

\begin{theorem}[Characterization] \label{thm:Barc}
  A non-failed Boolean CSP is closed under the applications of the
  rules of the proof system {\it BOOL\/} iff it is hyper-arc
  consistent.
\end{theorem}

\Proof Let $\phi$ be the CSP under consideration.
Below $C := x \A y = z$ is some {\em AND\/} constraint belonging to $\phi$.
We view it as a constraint on the variables $x,y,z$. Let
$D_x, D_y$ and $D_z$ be respectively the domains of $x, y$ and $z$. 

\NI
($\Ra$)
Consider the {\em AND\/} constraint $C$.
We have to analyze six cases.
\II

\NI
{\em Case 1.} Suppose $1 \in D_x$. 

Assume that neither
$(1,1) \in D_y \times D_z$ nor
$(0,0) \in D_y \times D_z$. Then either
$D_y = \C{1}$ and $D_z = \C{0}$ or
$D_y = \C{0}$ and $D_z = \C{1}$. 

If the former holds, then by the {\em AND 3\/} rule 
we get $D_x = \C{0}$ which is a contradiction.
If the latter holds, then by the {\em AND 5\/} rule
we get $D_z = \C{0}$ which is a contradiction.

We conclude that for some $d$ we have $(1,d,d) \in C$.
\II

\NI {\em Case 2.} Suppose $0 \in D_x$. 

Assume that $0 \not\in D_z$. Then  $D_z = \C{1}$, so
by the {\em AND 6\/} rule we get $D_x = \C{1}$ which is
a contradiction. Hence $0 \in D_z$. Let now $d$ be some element 
of $D_y$. We then have $(0,d,0) \in C$. 
\II

\NI
{\em Case 3.} Suppose $1 \in D_y$. 

This case is symmetric to Case 1.
\II

\NI {\em Case 4.} Suppose $0 \in D_y$. 

This case is symmetric to Case 2.
\II

\NI
{\em Case 5.} Suppose $1 \in D_z$. 

Assume that $(1,1) \not\in D_x \times D_y$.
Then either $D_x = \C{0}$ or $D_y = \C{0}$.
If the former holds, then by 
the {\em AND 4\/} rule we conclude that $D_z = \C{0}$.
If the latter holds, then by 
the {\em AND 5\/} rule we conclude that $D_z = \C{0}$.
For both possibilities we reached a contradiction. So 
both $1 \in D_x$ and $1 \in D_y$ and consequently $(1,1,1) \in C$.
\II

\NI {\em Case 6.} Suppose $0 \in D_z$. 

Assume that both $D_x = \C{1}$ and $D_y = \C{1}$.
By the {\em AND 1\/} rule we conclude that $D_z = \C{1}$ which
is a contradiction. So 
either $0 \in D_x$ or $0 \in D_y$ and consequently for some $d$
either $(0,d,0) \in C$ or $(d,0,0) \in C$.
\II

\NI
($\La$)
We need to consider each rule in turn.
We analyse here only the {\em AND\/} rules. For other rules the
reasoning is similar.
\II

\NI
{\em AND 1\/} rule.

Suppose that $D_x = \C{1}$ and $D_y = \C{1}$. 
If $0 \in D_z$, then by the hyper-arc consistency for 
some $d_1 \in D_x$ and $d_2 \in D_y$ we have $(d_1,d_2,0) \in C$, 
so $(1,1,0) \in C$ which is a contradiction. 

This shows that $D_z = \C{1}$ which means that 
$\phi$ is closed under the applications of this rule.
\II

\NI
{\em AND 2\/} rule.

Suppose that $D_x = \C{1}$ and $D_z = \C{0}$. 
If $1 \in D_y$, then by the hyper-arc consistency for 
some $d_1 \in D_x$ and $d_2 \in D_z$ we have $(d_1,1,d_2) \in C$, 
so $(1,1,0) \in C$ which is a contradiction. 

This shows that $D_y = \C{0}$ which means that 
$\phi$ is closed under the applications of this rule.
\II

\NI
{\em AND 3\/} rule.

This case is symmetric to that of the {\em AND 2} rule.
\II

\NI
{\em AND 4\/} rule.

Suppose that $D_x = \C{0}$. 
If $1 \in D_z$, then by the hyper-arc consistency for 
some $d_1 \in D_x$ and $d_2 \in D_y$ we have $(d_1,d_2,1) \in C$, 
so $(1,1,1) \in C$ which is a contradiction. 

This shows that $D_z = \C{0}$ which means that 
$\phi$ is closed under the applications of this rule.

\II

\NI
{\em AND 5\/} rule.

This case is symmetric to that of the {\em AND 4} rule.
\II

\NI
{\em AND 6\/} rule.

Suppose that $D_z = \C{1}$. 
If $0 \in D_x$, then by the hyper-arc consistency for 
some $d_1 \in D_y$ and $d_2 \in D_z$ we have $(0,d_1,d_2) \in C$, 
so $0 \in D_z$ which is a contradiction. 

This shows that $D_x = \C{1}$. By a symmetric argument also $D_y = \C{1}$
holds. 
This means that $\phi$ is closed under the applications of this rule.
\II

An analogous reasoning can be spelled out
for the equality, {\em OR\/} and {\em NOT\/} constraints and is omitted.

\HB
\III

Note that the restriction to non-failed CSP's is necessary:
the failed CSP $\p{x \A y = z}{x \in \ES, y \in \C{0,1}, z \in \C{0,1}}$
is not hyper-arc consistent but it is closed under the
applications of the rules of {\it BOOL}.

It is also easy to check that all the rules of the {\it BOOL\/} system
are needed, that is, this result does not hold when any of these 20
rules is omitted.  For example, if rule {\em AND 4\/} is left out,
then the CSP $\p{x \A y = z}{x = 0, y \in \C{0,1}, z \in \C{0,1}}$ is
closed under the applications of all remaining rules but is not
hyper-arc consistent.

In view of the fact that all considered proof rules preserve
equivalence, the above theorem shows that to reduce a Boolean
CSP to an equivalent one that is either failed or hyper-arc consistent
it suffices to close it under the applications of the rules of the
{\it BOOL\/} system.  This provides a straightforward algorithm for
enforcing hyper-arc consistency for Boolean constraints. We shall
return to this point in the final section.

\section{The Proof System of Codognet and Diaz}
\label{sec:cd}

Usually, slightly different proof rules are introduced when dealing
with Boolean constraints.  For example, in \citeasnoun{CD96} the set
of rules given in Table \ref{tab:bool'} is considered. We call the
resulting proof system {\em BOOL'}.

\begin{table}[htbp]
    \leavevmode
\[
\begin{array}{|ll|}  \hline
{\it EQU \ 1-4} & \mbox{ as in the system {\it BOOL}} \\[2mm]
{\it NOT \ 1-4} & \mbox{ as in the system {\it BOOL}} \\[2mm]
{\it AND \ 1'} & x \A y = z, x = 1 \ra y = z \\
{\it AND \ 2'} & x \A y = z, y = 1 \ra x = z \\
{\it AND \ 3'} & x \A y = z, z = 1 \ra x = 1 \\
{\it AND \ 4} & \mbox{ as in the system {\it BOOL}} \\ 
{\it AND \ 5} & \mbox{ as in the system {\it BOOL}} \\ 
{\it AND \ 6'} & x \A y = z, z = 1 \ra y = 1 \\[2mm]
{\it OR \ 1}  & \mbox{ as in the system {\it BOOL}} \\ 
{\it OR \ 2'}  & x \Or y = z, x = 0 \ra y = z \\
{\it OR \ 3'}  & x \Or y = z, y = 0 \ra x = z \\
{\it OR \ 4'}  & x \Or y = z, z = 0 \ra x = 0 \\
{\it OR \ 5}  & \mbox{ as in the system {\it BOOL}} \\ 
{\it OR \ 6'}  & x \Or y = z, z = 0 \ra y = 0 \\ \hline
\end{array}
\]
    \caption{Proof system {\em BOOL'}}
    \label{tab:bool'}
\end{table}

To be precise, the rules {\em EQU 1--4} are not present in \citeasnoun{CD96}.
Instead, the constraints $0=0$ and $1=1$ are adopted as axioms.
Note that rules {\em AND 1', AND 2', OR 2'\/} and  {\em  OR 3'\/}
introduce constraints in their conclusions.
{\em OR 2'\/} rule corresponds to rule (\ref{equ:rule2})
of Section \ref{subsec:motivation}.

The main difference between {\it BOOL} and {\it BOOL'} lies in the
fact that the rules {\em AND 1--3\/} of {\it BOOL} are replaced by
the rules {\em AND 1'\/} and {\em AND 2'\/} of {\it BOOL'}
and the rules {\em OR 2--4\/} of {\it BOOL} are replaced by
the rules {\em OR 2'\/} and {\em OR 3'\/} of {\it BOOL'}.
(The fact that the rule {\em AND 6\/} of {\it BOOL} is split in {\it BOOL'}
into two rules, {\em AND 3'\/} and {\em AND 6'\/}
and analogously for the rules {\em OR 6 \/} of {\it BOOL}
and  {\em OR 3'\/} and {\em OR 6'\/} of {\it BOOL'}
is of no importance.)

The {\em   AND\/} rules of the  {\it BOOL'\/} system
can be found (in a somewhat different format) in \citeasnoun{simonis89a}.
A natural question arises whether the proof systems {\it BOOL} and
{\it BOOL'} are equivalent. The precise answer is ``sometimes''.
First, observe that the following result holds.

\begin{theorem} \label{thm:Barc1}
If a non-failed Boolean CSP is closed under the
applications of the rules of the proof system {\it BOOL'},
then it is hyper-arc consistent.
\end{theorem}
\Proof
The proof relies on the following immediate observation.
\II

\NI
{\bf Claim}
{\em
Consider a Boolean CSP $\phi$ containing the {\it AND\/} constraint $x \A y = z$
on the variables $x,y,z$ with respective domains
$D_x, D_y$ and $D_z$. 
If  $\phi$ is closed under the
applications of the {\it AND 1'} rule, then $D_x = \C{1}$ implies $D_y = D_z$.
If  $\phi$ is closed under the
applications of the {\it AND 2'\/} rule, then
$D_y = \C{1}$ implies $D_x = D_z$.}
\HB
\II

Suppose now that the CSP in question contains the {\it AND\/}
constraint $x \A y = z$ on the variables $x,y,z$ with respective domains
$D_x, D_y$ and $D_z$. We present the proof only for the cases
where the argument differs from the one given in the proof
of the hyper-arc consistency Theorem \ref{thm:Barc}.
\II

\NI
{\em Case 1.} Suppose $1 \in D_x$. 

Assume that neither
$(1,1) \in D_y \times D_z$ nor
$(0,0) \in D_y \times D_z$. Then either
$D_y = \C{1}$ and $D_z = \C{0}$ or
$D_y = \C{0}$ and $D_z = \C{1}$. 

If the former holds, then by Claim 1 $D_y = D_z$, which is a contradiction.
If the latter holds, then by the {\em AND 5\/} rule $D_z = \C{0}$ which
is also a contradiction.
We conclude that for some $d$ we have $(1,d,d) \in C$.
\II

\NI {\em Case 6.} Suppose $0 \in D_z$. 

Assume that both $D_x = \C{1}$ and $D_y = \C{1}$.
By Claim 1 $D_y = D_z$, which is a contradiction. So
either $0 \in D_x$ or $0 \in D_y$ and consequently for some $d$
either $(0,d,0) \in C$ or $(d,0,0) \in C$.
\II

The reasoning for other Boolean constraints is analogous and omitted.
\HB 
\III

In contrast to the case of the {\it BOOL} system the converse 
result does not hold.
Indeed, just take the CSP
$\phi := \p{x \A y = z}{x=1, y \in \C{0,1}, z \in \C{0,1}}$.
Note that $\phi$ is hyper-arc consistent but it is not closed under the
applications of the {\em AND 1'\/} rule.

In general, there are four such ``problematic'' CSP's.  In each of
them the single {\em AND \/} or {\em OR \/} constraint can be
reduced to an equality constraint. These four CSP's are used in the
following definition.

\begin{definition}
We call a  Boolean CSP {\em limited\/} if none of the following four
CSP's forms a subpart of it:
\begin{itemize}
\item $\p{x \A y = z}{x=1, y \in \C{0,1}, z \in \C{0,1}}$,
\item $\p{x \A y = z}{x \in \C{0,1}, y=1, z \in \C{0,1}}$,
\item $\p{x \Or y = z}{x=0, y \in \C{0,1}, z \in \C{0,1}}$,
\item $\p{x \Or y = z}{x \in \C{0,1}, y=0, z \in \C{0,1}}$.
\HB
\end{itemize}
\end{definition}

The idea is that if we exclude these ``problematic'' CSP, then
hopefully we prevent the situation that a CSP is 
hyper-arc consistent but is not closed under the
applications of the {\em AND 1'} (respectively
{\em AND 2', OR 2'\/} or {\em OR 3'}) rule.
This is exactly what the following theorem states.

\begin{theorem} \label{thm:Barc2}
If a non-failed Boolean CSP is limited and hyper-arc consistent,
then it is closed under the
applications of the rules of the proof system {\it BOOL'}.
\end{theorem}
\Proof
In view of the hyper-arc consistency Theorem \ref{thm:Barc} we only have
to consider the rules of {\it BOOL'} that are absent
in {\it BOOL}. We present here an argument for one representative
rule.
\II

\NI
{\em AND 1'} rule.

Suppose that $D_x = \C{1}$. 
If $0 \in D_y$, then by the hyper-arc consistency for some $d \in D_z$ we have
$(1,0,d) \in C$, which means that $0 \in D_z$.
Conversely, if $0 \in D_z$, then by the hyper-arc consistency for 
some $d \in D_y$ we have $(1,d,0) \in C$, so $0 \in D_y$.
By a similar argument we get that $1 \in D_y$ iff $1 \in D_z$.
This shows that $D_y = D_z$.

By assumption $\phi$ is limited, so either $D_y \neq \C{0,1}$ or $D_z
\neq \C{0,1}$. Hence either $D_y = D_z = \C{1}$ or $D_y = D_z =
\C{0}$. In both cases the CSP under consideration is closed under the
applications of the {\em AND 1'} rule.  
\HB
\III

To summarize: for Boolean CSP's that are limited the respective
closures under the rules of the proof systems {\it BOOL\/} and {\it
BOOL'\/} coincide.

\section{Relation to the {\tt CHR} Language}

The rules such as the ones given in the proof system {\it BOOL\/} can
be straightforwardly represented as so-called simplification rules of
the {\tt CHR} language of \citeasnoun{fruhwirth-constraint-95}.  The
{\tt CHR} language is part of the \eclipse{} system (see
\citeasnoun{Ecl95}).  For a more recent and more complete overview of
{\tt CHR} see \citeasnoun{FruehwirthJLP98}.  For example {\em AND 6}
rule, so
\[
x \A y = z, z = 1 \ra x = 1, y = 1,
\]
is written in the syntax of {\tt CHR} as
\[
{\tt and(X,Y,Z) <=> Z = 1 \ | \ X = 1, Y = 1}.
\]

In fact, such {\tt CHR} rules for the {\em AND\/} constraint can be
already found in \citeasnoun{fruhwirth-informal}.  They amount to the
corresponding {\em AND\/} rules of the {\em BOOL'} system.
Boolean constraints form a prime example for an effective use of {\sf
  CHR}s.  A {\tt CHR} program that corresponds to the proof system
{\it BOOL\/} or {\it BOOL'\/} when combined with a labeling procedure
constitutes a natural decision procedure for Boolean CSP's.  The
Characterization Theorem \ref{thm:Barc} shows that the {\tt CHR} rules
corresponding to the {\it BOOL\/} system implement hyper-arc
consistency.

\section{Conclusions}

In this paper we collected a number of simple but hopefully useful
observations on Boolean constraint propagation rules.  First of all,
we clarified in what sense one set of such rules is complete. Then we
showed that Boolean constraint propagation is in fact equivalent to
unit propagation, a form of resolution for propositional logic. The
reduction in each direction can be achieved in constant time.

This shows that given a combinatorial problem that can be
naturally formalized using Boolean constraints (for example, a problem
concerning combinatorial circuits) it is useless to translate it to a
clausal form and subsequently employ unit propagation: in such case
Boolean constraint propagation achieves the same effect.  Conversely,
it is useless to translate a clausal form representation to a
representation that uses Boolean constraints with the aim of employing
Boolean constraint propagation: in this case unit propagation achieves
the same effect.

The subsequent characterization of the introduced set of Boolean
constraint propagation rules by means of the hyper-arc consistency
notion shows that this set of rules is in some sense optimal.  The
notion of hyper-arc consistency also allowed us to differentiate
between two sets of such rules proposed in the literature.

\section*{Acknowledgement}
We thank Rina Dechter, Thom Fr\"{u}hwirth and the referees 
for helpful comments.



\end{document}